\documentclass[journal,10pt]{IEEEtran}
\usepackage{amsmath,amsfonts}
\usepackage{algorithmic}
\usepackage{algorithm}
\usepackage{array}
\usepackage[caption=false,font=normalsize,labelfont=sf,textfont=sf]{subfig}
\usepackage{textcomp}
\usepackage{stfloats}
\usepackage{url}
\usepackage{verbatim}
\usepackage{graphicx}
\usepackage{cite}
\usepackage{authblk}
\usepackage{setspace}
\usepackage{booktabs}
\usepackage{pdflscape}
\usepackage[utf8]{inputenc}
\usepackage{mathtools}
\usepackage{xspace}
\usepackage{rotating}
\usepackage{xr-hyper}
\usepackage{dirtree}
\usepackage{mathtools}
\usepackage{xcolor}
\usepackage{rotating}
\usepackage{longtable}

\date{}

\def\masc{\mbox{\emph{MaSC}}\xspace}
\def\mascs{\mbox{\emph{MaSC}}'s\xspace}

\def\ie{\emph{i.e.}\xspace}

\def\yolo4{\textsc{YOLO}v4\xspace}

\def\yolon{\textsc{YOLO}v9\xspace}

\def\yolo{\textsc{YOLO}\xspace}
\def\cnv2{\emph{CorNetv2}\xspace}
\def\cnvt{\emph{DroneZaic}\xspace}
\def\cnvtt{\emph{CorNetv3}\xspace}

\def\smd{\emph{MSDD}\xspace}
\def\cn{\mbox{\emph{CorNet}}\xspace}
\def\asift{\textsc{ASIFT}\xspace}	  
\def\uav{\textsc{UAV}\xspace}  
\def\uavs{\textsc{UAV}s\xspace}

\def\mascf{\mbox{\emph{MaizeStandCounting}}\xspace}
\def\dmc{\mbox{\emph{MaSC}}\xspace}

\def\exg{\mbox{ExG}\xspace}
\def\yv12{\mbox{YOLOv12}\xspace}
\def\yolo{\mbox{YOLO}\xspace}
\def\yvn{\mbox{YOLOv9}\xspace}

\def\webodm{\text{WebODM}\xspace}

\usepackage{hyperref}

\begin{document}

\title{\mascf (\masc): Automated and Accurate Maize Stand Counting from UAV Imagery Using Image Processing and Deep Learning}

\author{Dewi Endah Kharismawati,~\IEEEmembership{Member,~IEEE,}
        and Toni Kazic

\thanks{D. E. Kharismawati and T. Kazic are with the Department of Electrical Engineering and Computer Science, University of Missouri, Columbia, USA (e-mail: dek8v5@missouri.edu; kazict@missouri.edu).}
\thanks{Manuscript received 08 October 2025;}
}

\markboth{Journal of IEEE Journal of Selected Topics in Signal Processing, 08 October 2025}%
{Shell \MakeLowercase{\textit{et al.}}: A Sample Article Using IEEEtran.cls for IEEE Journals}



\maketitle

\begin{abstract}
Accurate maize stand counts are vital for both crop management and
agricultural research, as they inform key decisions related to yield
prediction, planting density optimization, and early identification of
germination issues.
In commercial settings, stand counts help farmers determine the need for
replanting, assess planting equipment performance, and guide input
applications such as fertilizers and herbicides.
In research contexts, they are critical for comparing genetic lines,
evaluating treatment effects, and maintaining experimental consistency.
However, traditional manual counting methods, which involve walking through
fields and counting plants row by row, are labor-intensive, time-consuming,
and prone to human error, especially across large or heterogeneous plots.
These limitations demand efficient, accurate, and scalable
automated solutions.
To address this challenge, we present \mascf (\masc), a robust, end-to-end
algorithm for automated maize seedling stand counting using RGB imagery
captured by low-cost UAVs and processed on affordable computing
hardware.
Depending on the input images, \masc operates in two modes: (1) mosaic
images through patchification or (2) raw video frames with associated homography
matrices.
Both modes employ a lightweight YOLOv9 model trained to detect maize
seedlings at the V2--V10 growth stages.
\masc accurately distinguishes maize from weeds and other vegetation and
performs row and range segmentation based on the spatial distribution of
detected plants, enabling precise row-wise stand counts.
We evaluated \masc against in-field manual counts across diverse maize lines
in our 2024 summer nursery.
The algorithm achieved an R² of 0.616 using mosaics and 0.906 using raw
frames, demonstrating strong agreement with ground truth data.
Additionally, \masc processed 83 full-resolution frames in just 60.63
seconds, including both model inference and post-processing, highlighting
its potential for real-time performance and deployment on farming tractors
or onboard \uavs.
These results underscore \mascs potential as a scalable, low-cost, and
accurate tool for automated maize stand counting in both research and
production settings.
\end{abstract}

\section{Introduction}
\label{sec:org91b8639}

\label{intro}

Localizing, identifying, and tracking plants during the growing season are
essential to characterizing a wide variety of phenotypes in genetic
experiments.
Similarly, optimizing growing conditions for production crops
relies heavily on detecting plant growth and stresses.
Stand counts --- determining the number of plants of interest in each row
or field --- are the first step in a season-long process of collecting
developmental and morphological phenotypes, beginning with germination and
lethality.
As the growing season progresses and phenotypes are expressed, the ability
to minimize effort by dynamically targeting scouting operations becomes
increasingly important.
Today, geneticists commonly count their plants by walking the field and
counting the plants in each row.
This method is labor intensive, slow, and prone to human error.
Automating stand counts is the obvious solution, but the current equipment
entails significant trade-offs, contributing to the phenotyping bottleneck
\cite{bullock1998,furbank2011,khaki2020}.
Well-equipped farmers with GPS tractors can pass through their fields until
the plants are too high, preventing further scouting as canopy closure occurs.
Smaller aerial and ground-based sensors have been developed to permit
scouting throughout the growing season.
For example, \cite{kayacan2018} developed an ultra compact, 3D-printed
field robot for scouting maize fields.
These can produce very accurate counts at different growth stages, but many
robots are larger than the space between rows or only image seedlings
\cite{fan2022,debruin2025,zhu2022,seespray2025}.
Ground-based robots require a large amount of energy to travel the uneven
surfaces of fields, and using fixed, grid-search trajectories limit
coverage area and increase the number missions needed
\cite{kayacan2018,zhang2020,wang2023b,shafiekhani2017}.
While ground-based robots can achieve high accuracy, their size and
operational limitations hinder scalability and affordability. In contrast,
unmanned aerial vehicles (\uavs) offer broad coverage with less physical
interference, encouraging their development for high-throughput phenotyping
\cite{djiag2023,parrotgrass2018,xag2025}.
Most vehicles in current use can fly about 25 minutes on a single battery,
and the batteries are small and light enough to easily bring several to
complete an afternoon's data collection.
However, most of these vehicles are expensive and engineering-intensive in
order to carry multiple sensors and supply the positional metadata many
algorithms require.
These costs preclude widespread adoption of these vehicles, especially for
academic researchers and small stakeholders.
Even well-equipped groups can find it difficult to schedule robotic surveys
frequently enough to pinpoint the onset of phenotypes (F. Fritschi,
personal communication), and fixed trajectories can entail an
inconveniently large number of missions for little information gain.
Given data, transforming imagery into actionable phenotypic information
remains a major challenge.
The physical traits of plants can be observed with modern
sensors that collect voluminous data, and machine learning and image
processing are rapidly advancing
\cite{minervini2015,furbank2011,kharismawati2022,nabwire2021,li2020a,feng2020,ndubuisi2023a,fahlgren2015a}.
Nonetheless, even basic challenges such as distinguishing plants of
interest from weeds remain, especially during early growth stages when
visual similarities are pronounced \cite{wang2023b,mahlein2016}.
Many approaches rely on detecting green in the imagery to compute the
vegetation index \cite{mardanisamani2022}.
This is computationally easy and cheap, but confuses the plants of interest
with the others that can occur in a row, such as weeds and grasses
\cite{woebbecke1995,meyer2004,rouse1974}.
More sophisticated approaches to identifying maize involve segmenting
vegetation using the excess greenness (\exg) index, followed by the
extraction of geometric features. These features are then used in a
decision tree classifier to distinguish maize plants from other green
elements, such as weeds or overlapping vegetation \cite{varela2018}.
A fundamental challenge to this strategy is building a sufficient
ground-truth data set: genetic experiments can involve hundreds of
different lines, each with different developmental and morphological
phenotypes, planted at several different times.
Imaging the plants from different positions and camera poses and altitudes
introduces additional variation.
An alternative is simply to train a neural network for stand counting.
Kitano et al. \cite{kitano2019} used U-Net for pixel-wise segmentation,
followed by morphological opening and blob detection to count individual
plants.
However, the approach was sensitive to UAV altitude, causing
failures when seedlings appeared too small.
Katari et al. \cite{katari2024} proposed an automated labeling and
CNN-based framework for plant counting.
While effective for stand estimation, their method identifies plant regions
only approximately, making it unsuitable for fine-grained phenotyping tasks
that require pixel-level accuracy.
Wang et al. \cite{wang2023b} applied \yolo detectors on orthomosaics to estimate
stand counts and spacing variability.
While effective, their reliance on
mosaicked imagery and bounding-box localization introduces artifacts and
limits precision, particularly under overlapping canopies.

To address these challenges, we present a computational system that
provides accurate stand counts for research and production fields using
simple, low-cost \uavs and processing hardware.
A system compact and
efficient enough to operate on consumer-grade devices in real time would
enable dynamic scouting of large areas and targeted inspection of specific
regions of interest.
Our initial experiments with traditional image processing methods ---
including segmentation using \exg, watershed, and distance transform to
generate Voronoi cells for individual plants --- demonstrated their
limitations.
These techniques struggled to separate overlapping plants and
were unable to reliably distinguish maize from other green objects, as \exg
is solely color-dependent.
We propose \mascf (\masc), an end-to-end stand counting pipeline that processes
UAV imagery and outputs stand counts for each row.
\masc accepts two types of input: (1) image mosaics and (2) raw videos
captured by \uavs.
To preprocess video input, we employ our custom mosaicking algorithms to
generate a complete view of the field while preserving high resolution
\cite{aktar2018,aktar2020,kharismawati2020,kharismawati2025a}.
Accurate image mosaicking helps prevent over- or under-counting by
summarizing a sequence of frames into a single composite image, ensuring
that each plant is represented only once.
In mosaic mode, the full mosaic is divided into patches that match the
input size of the detection model, and the resulting bounding boxes are
later stitched together based on patch coordinates.
In raw video mode, individual frames are passed directly to the detection
model, and the homography matrices computed during mosaicking are used to
register both frames and their bounding boxes.
For detection, we use \yolo, a supervised convolutional neural network that
segments and identifies objects.
\yolo is widely used for its combination of high speed, detection accuracy,
and continual improvements driven by the open-source research community
\cite{redmon2018,bochkovskiy2020,wang2022,wang2024,jocher2023,tian2025}.
We trained \yolo to recognize seedling maize plants from the V2--V12
growth stages (approximately 10–100 cm in height), using our publicly
available dataset, introduced and benchmarked in \cite{kharismawati2025e}.
The dataset contains images of seedlings appearing singly or in groups of
two and three. 
We evaluated \masc against in-field, manual stand counts across several
maize lines in both nursery and production field conditions.
\masc achieved an \(R^2\) of 0.616 when detecting from patchified
mosaics, while detection directly from raw frames yielded a significantly
higher \(R^2\) of 0.906.
\masc can process 83 full-resolution frames in just 50.63 seconds, with
detection taking only 25.02 seconds, highlighting its computational
efficiency.
These results demonstrate that \masc is not only accurate and fast, but also
lightweight enough to be deployed on real-world agricultural platforms such
as UAVs or tractors.
By reducing manual labor and enabling scalable, in-field stand counting,
\masc offers a practical tool for both researchers and producers seeking
timely crop monitoring.

\section{Materials and Methods}
\label{sec:org024fc95}

\label{mnms}

\label{mats}

\subsection{Maize Nurseries, Video Collection, and Computational Equipment}
\label{sec:orgaff2c25}

Maize genetic nurseries were planted and imaged in 2019 and 2024, as
described in \cite{kharismawati2025e}.
Our fields were planted either by hand using a jab planter for the disease
lesion mimic mutant and inbred lines, with a Jang rotary push planter for
the elite lines, or by machine along the borders.
Rows are 6.1 meters long and run east–west, with 0.91-meter spacing between
rows.
A set of rows running north–south is referred to as a range.
In the imaged hand-planted fields, ranges are separated by 1.22-meter
unplanted alleys, though row length and plant spacing vary by
investigator.
Machine-planted fields do not include alleys.
We describe the growth stages of the maize plants using the standard
``leaf collar'' method, which counts the number of visible leaves, starting
at 1 for the coleoptile \cite{nielsen2004}.
For our lines, the approximate
average height is 10 cm at V4, 50 cm at V8, 70 cm at V10, and 100 cm at V12.
RGB video imaging was performed using the
\href{https://www.dji.com/phantom-4}{DJI Phantom 4 Pro}
and
\href{https://www.dji.com/mavic-2}{DJI Mavic 2 Pro}
drones, manufactured by Da-Jiang Innovations, Shenzhen, China.
All flights were flown manually using the Autopilot mobile app
(\href{https://autoflight.hangar.com/autopilot/flightschool}{Autopilot})
and DJI GO 4.
Flight paths were mainly serpentine, running both parallel
and perpendicular to the planting rows.
All imaging took place at relatively low altitudes, approximately 5--30
meters above ground level (AGL).
Flights were conducted in light winds, though occasional horizontal
and vertical movement occurred due to air currents.
\uav speed varied, but remained below 3.2 kilometers per hour. Video was
collected using both nadir and oblique camera angles at 24 and 30 frames
per second (fps) in 24-bit color depth.
All code was run on a Lambda Labs machine with an Intel Core i9-9920X CPU,
two NVIDIA RTX 2080Ti GPUs, and 128 GB of memory.

\subsection{Image Data Sets}
\label{sec:orgef477c5}
We present three seedling image datasets: one captured in 2019 for an
initial small-scale color segmentation experiment, one from 2020 covering
full-field color segmentation, and a third collected in 2024 for deep
learning-based detection.
The second dataset was captured on July 2nd, 2020.
The flight followed a simple, single forward pass at an altitude of 30
meters AGL.
To achieve higher resolution and broader coverage, the second data set was
captured on June 12, 2024, using a more complex flight pattern.
Flights were conducted at 10 meters AGL, with trajectories oriented perpendicular
to the planting rows and 70\% overlap between adjacent ranges at each pass.
Flight paths alternated between the center of the crop rows and the center
of the alleys.

\subsection{Color-Based Corn Segmentation}
\label{sec:org2962206}

\begin{figure}
\centering
\includegraphics[width=1\linewidth]{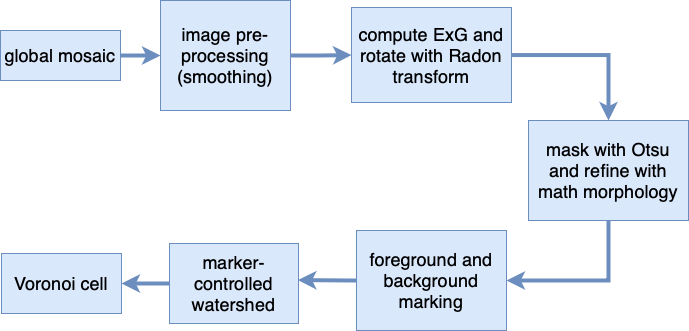}
\caption{Workflow of segmentation with color.}
\label{exg-workflow}
\end{figure}

The color-based segmentation workflow is shown in Figure
\ref{exg-workflow}.
The process begins by applying a median filter to the original input mosaic
image to reduce noise while preserving structural boundaries (Eq. \ref{med-fil-eqn}).
\begin{equation} 
\hat{f}(x,y) = \mathrm{median}\{g(s,t)\},  (s,t) \in S_{xy},
\label{med-fil-eqn} 
\end{equation} 
where \(g(s,t)\) is the pixel value at \((s,t)\), and \(S_{xy}\) is an \(m
\times n\) window centered at pixel \((x,y)\). This operation enhances edge
definition between vegetation and soil while removing small outliers
\cite{ariascastro2009,narendra1981}.
Then, the Excess Green Index
(\exg) is computed for each pixel in the mosaic image to distinguish plants from the
background, using Equation \ref{exg-eq}:
\begin{equation}
\text{\exg} = 2 \times \text{Green} - \text{Red} -  \text{Blue},
\label{exg-eq}
\end{equation}
where \textit{Red}, \textit{Green}, and \textit{Blue} represent the pixel
intensity values of the respective RGB channels.
We use the \exg image to compute the Radon transform, and then take the
variance of the transform to quantify the dominant orientation of
vegetation rows.
This estimated angle is then used to rotate all images so
that the row orientation is standardized across different datasets,
aligning the vegetation rows from top to bottom.

A binary vegetation mask is then obtained by applying Otsu's thresholding
method to the filtered \exg image, automatically determining the threshold
that best separates plant and non-plant regions \cite{otsu1979}.
These preliminary masks often have angular contours, both at the tips of
leaves and where they curve downward toward the soil.
Additionally, leaves
that touch those of adjacent plants (``touching corn'') can result in merged
regions that group multiple plants together in morphologically incorrect
ways.
To address this, the masks are refined using morphological erosion
operations \cite{serra1983}, with a disk-shaped structuring element. The
element's size is computed automatically from the average minor axis length
of the connected components in the mask, avoiding manual parameter
selection.

Objects were segmented and localized using the Euclidean distance transform
followed by watershed segmentation, a technique widely used in biomedical
image analysis to separate clustered objects
\cite{abdolhoseini2019,yang2006,disttrans2019}.
The binarized vegetation mask is first processed using an \(L_2\) (Euclidean)
distance transform, which computes, for each foreground pixel, its distance
to the nearest background pixel.
The resulting distance map is normalized to the range [0,1],
producing smooth gradients where the centers of plant regions correspond to
local maxima.
These regional maxima are used as internal markers: ideally,
each plant produces a single maximum, while clusters of touching plants
yield multiple peaks \cite{disttrans2019,liu2010a}.
These markers are then used to initialize the watershed algorithm, which
floods the distance map from each marker outward.
The algorithm segments
the image by identifying the ridge lines where these growing regions meet,
thereby separating adjacent or overlapping plants and localizing each plant
instance.

\subsection{Input Handling for \masc Workflow}
\label{sec:org8651b98}

\begin{figure}[hbt]
\centering
\includegraphics[width=1\linewidth]{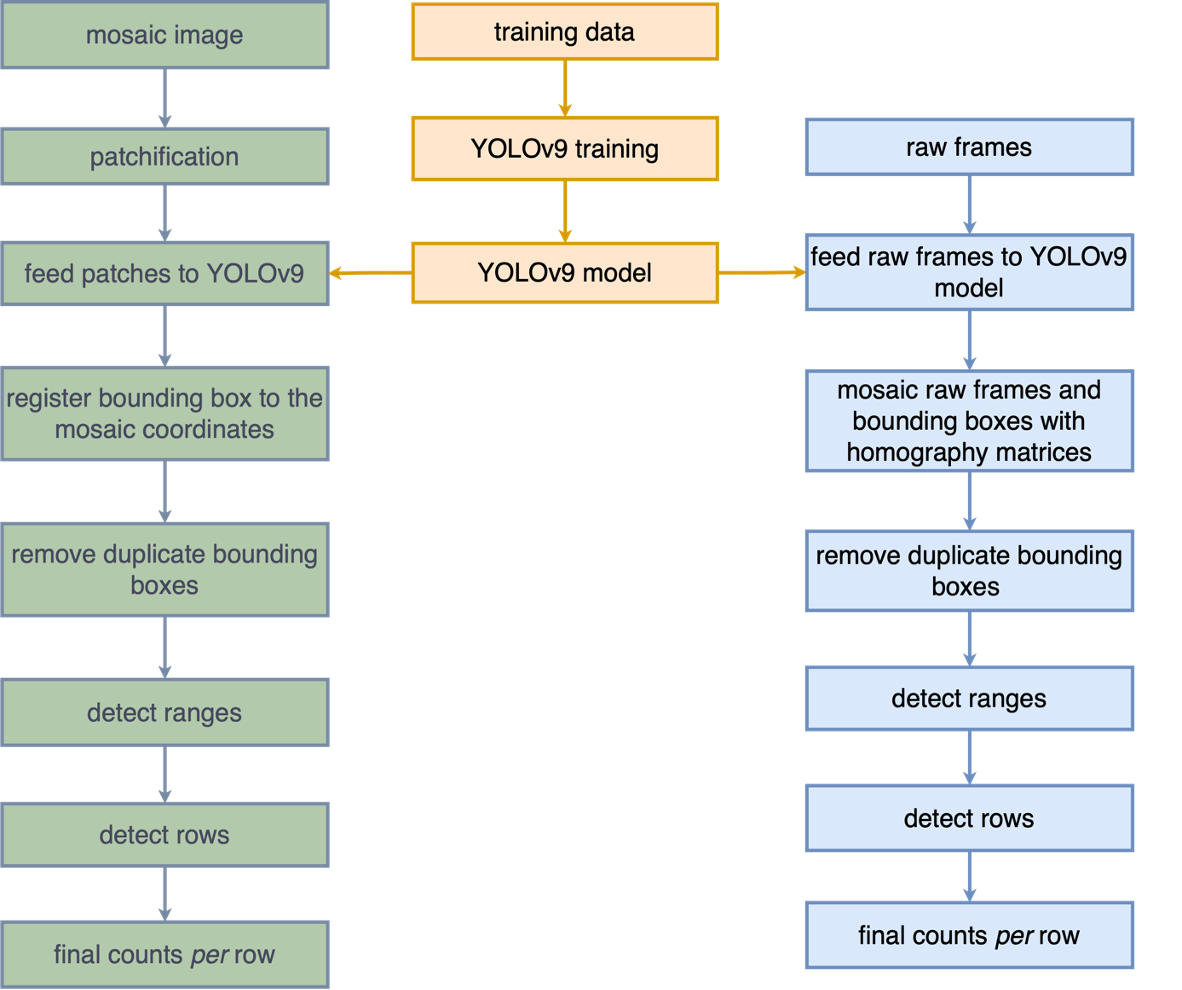}
\caption{Overview of the \masc processing workflow.
Green boxes represent the pipeline for pre-mosaicked image inputs.
The blue box indicates the raw video input mode, which includes internal mosaicking.
Orange boxes show the seedling detection process using \yvn, shared across both input modes.}
\label{dmc-complete-workflow}
\end{figure}

At the seedling stage, plants are very small, requiring high-resolution
imaging to accurately identify individual seedlings.
We obtain the required resolution by flying the \uav at very low altitude
(10 meter AGL in our 2024 trials), but this produces a
narrow field of view.
To recover the whole field of view of the field, the individual frames must
be stitched into a seamless mosaic so that: every plant has a unique
coordinate, and duplicate detections in overlapping frames can be
summarized.
\masc supports two complementary input modes (Figure \ref{dmc-complete-workflow}).

\subsubsection{Mode1: Mosaic Images}
\label{sec:org034a26d}

In the pre-mosaicked input mode, users supply a previously mosaicked image
generated using external software.
Typically, mosaicked images are large; for example, our 2024 test dataset
consists of images approximately \(5576 \times 11375\) pixels, captured at
altitudes between \(10\) and \(15\) meters AGL.
Such dimensions exceed the input size limitations of deep learning object
detection models like \yolo, which requires \(640\times640\) pixel input
images.
To manage this, large mosaicked images are subdivided into smaller patches
(patchification).
Users can define the patch size based on their seedling
resolution; however, we recommend using multiples of 640 pixels for optimal
compatibility.
During patchification, \masc records the coordinates of each
patch to facilitate accurate reassembly of detection results.
Additionally, users can specify an overlap percentage between patches to
minimize issues when patches split rows of seedlings.
The complete pre-mosaicked workflow is depicted by the green boxes in
Figure \ref{dmc-complete-workflow}.

\subsubsection{Mode2: Raw Frames}
\label{sec:org0649b55}

The second mode processes raw video captured directly by the \uav. In
this workflow, raw video undergoes complete internal processing using our
\cnvt pipeline, an accurate and efficient mosaicking algorithm detailed
in \cite{kharismawati2025a}.
\cnvt dynamically samples the raw video to
extract frames with uniform overlap, performs lens and gimbal calibration,
estimates homography matrices, conducts shot detection, and creates
mini-mosaics to reduce error accumulation.
For homography estimation, we use our deep learning-based methods, \cnvtt
and \cn, as well as \asift, a traditional feature descriptor that has
proven highly effective for agricultural imagery due to its superior
accuracy, despite being computationally intensive for seedling datasets.
The pipeline computes and stores pairwise homography matrices (\(3 \times
3\)) between successive frames, denoted as \(H_{i \rightarrow i+1}\), which
map coordinates from frame \(F_i\) to frame \(F_{i+1}\).
Each frame \(F_i\) is then projected onto the reference frame \(F_0\) using the
cumulative homography:
\begin{equation}
H_{0 \leftarrow i} = H_{0 \leftarrow 1} \times H_{1 \leftarrow 2} \times \cdots \times H_{i-1 \leftarrow i}
\end{equation}
where each \(H_{0 \leftarrow i}\) maps coordinates from frame \(F_i\) to frame \(F_0\).

Raw frames, typically sized at \(3840 \times 2160\) pixels, are directly
processed without additional patchification.
Each processed frame is passed to the \yvn seedling detection
model.
Detection results are stored in corresponding .txt files containing
\texttt{class\_id}, \texttt{centroid\_x}, \texttt{centroid\_y}, \texttt{width}, \texttt{height}, and \texttt{confidence\_value}
Due to overlapping frames, the same plant may be detected multiple
times.
These duplicate detections are resolved using Non-Maximum Suppression
(NMS), which selects the bounding box with the highest confidence score
while suppressing overlapping boxes with lower scores \cite{neubeck2006}.
This ensures accurate and consistent detection results in the final global
mosaic.
Non-Maximum Suppression is typically applied as:
\begin{equation}
\text{IoU}(B_i, B_j) = \frac{|B_i \cap B_j|}{|B_i \cup B_j|} > \tau \Rightarrow \text{Suppress } B_j
\label{nonmaxsup}
\end{equation}
where \(B_i\) and \(B_j\) are bounding boxes and \(\tau\) is the IoU
threshold for suppression.
In our implementation, we set \(\tau = 0.25\) (\ie, 25\%).
The raw video processing workflow is illustrated by the blue box in Figure
\ref{dmc-complete-workflow}.

\subsection{Training \yvn}
\label{sec:org2888363}

\begin{figure*}[ht]
\centering
\includegraphics[width=1\linewidth]{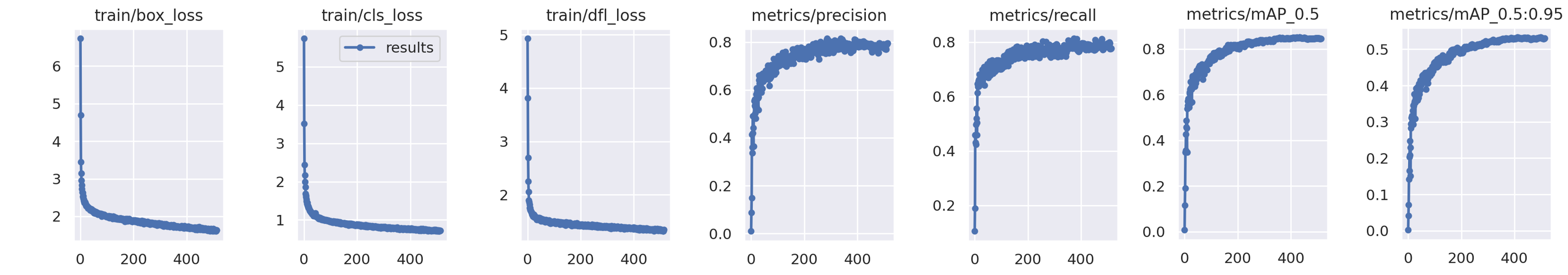}
\caption{
Training results of YOLOv9. The plots show the evolution of box loss,
classification loss, and distribution focal loss during training (left),
along with the corresponding performance metrics: precision, recall,
mean average precision at IoU=0.5 (mAP@0.5), and mean average precision at IoU=0.5:0.95
(mAP@0.5:0.95) (right).
}
\label{train-res}
\end{figure*}

The training, validation, and test dataset used in this paper to train the
\yvn model is publicly available and detailed in
\cite{kharismawati2025e}.
To improve accuracy, the dataset includes three classes: single, double,
and triple plants.
The number of double and triple instances is low due to a low rate of
planting errors.
Benchmarking results indicate that \yvn achieves the highest mean average
precision at an IoU threshold of 0.5 (mAP@0.5), as well as superior
precision and recall for single-plant detection, although it exhibits a
slower inference speed compared to YOLOv11.
Given that our current focus is on detection accuracy, this trade-off is
acceptable.
The training was conducted on our Lambda Labs machine.
Training was initialized using the yolo9c pre-trained
weights.
The chosen architecture incorporates cross-scale feature fusion,
dynamic label assignment, and a compound backbone, enhancing detection
robustness against variability in plant sizes and occlusions, as described
in \cite{wang2024}.
Prior to training, images underwent extensive preprocessing to enhance
model generalization capabilities. This preprocessing pipeline included
mosaic augmentation, resizing, and normalization, alongside additional
augmentations such as random horizontal flipping, brightness adjustments,
rotation, and scaling.
Training parameters included a batch size of 8, executed for a total of 513
epochs, resulting in a cumulative training duration of approximately 34
hours.
Figure \ref{train-res} shows the \yvn training progress summary plots.
All losses decrease rapidly and flatten out, while precision, recall, and
mAP steadily increase and plateau. The training appears stable and
effective, with no signs of overfitting or divergence.

\subsection{Range and Row Detection}
\label{sec:org8a5e20d}
\label{row-det}

\dmc provides different counting options for production and genetic nursery
fields.
In production fields, a single line of maize is planted in continuous rows,
without intervening unplanted alleys; thus, \dmc counts the entire length
of each row without any partitioning into ranges.
In genetic nurseries, however, different lines are planted in different
rows, and the rows are often separated by unplanted alleys  that divide
rows into separate ranges.
Row detection can be challenging due to uneven planting density, terrain
variability, row curvature over the length of the field, and mosaicking
quality.
Consequently, row coordinates determined in one range are not guaranteed
to translate accurately to other ranges.
Furthermore, the orientation of rows and ranges with respect to the image
or mosaic frame largely depends on the initial positioning and trajectory
of image acquisition.
For these reasons, nursery field stand counting proceeds by first detecting
the ranges, then rows within each range, and finally individual plants
within each row.
Range and row detection begins by standardizing row orientation
horizontally, using the Radon transform to estimate the angle at which rows
are rotated relative to the image frame \cite{pelapur2013}.
The Radon transform is applied to \exg values (Equation \ref{exg-eq}) for every
angle from 1\textdegree{} to 180\textdegree{}, producing line-integral profiles that
highlight linear structures.
For each angle, we calculate the variance of the Radon projection; high
variance indicates clear alternation between bright crop rows and darker
inter-row area.
The angle with the maximum variance is selected as the dominant crop-row
orientation.
This approach provides a robust, efficient
estimation of row orientation, typically running in well under one second
per tile on a standard CPU.
Then, \dmc uses the centroids of detected bounding boxes to determine
ranges by summing centroid positions based on image height, smoothing the
results with a convolutional moving window, identifying peaks, and
subsequently locating gaps between these peaks.
For row detection within each identified range, centroid positions are
similarly summed based on image width, and the same smoothing and peak
detection procedure is applied.

\subsection{Evaluation}
\label{sec:orgdd7c84a}

Manual ground-truth counting was conducted in the field by three
individuals independently counting seedling stands in silence while walking
along each row.
Upon reaching the end of each row, a consensus count was
determined.
If the counters disagreed on the number of seedlings, the
counting process was repeated until agreement was reached.
This manual counting aimed to capture total germination, including
seedlings that had germinated but subsequently died, adding complexity to
nursery field counting.

Ground truth data collection occurred on June 18 and 19, 2024.
For evaluation, automated counts from \dmc for each row were compared to
manual ground-truth counts using the coefficient of determination (\(R^2\)),
calculated as follows:
\begin{equation}
R^2 = 1 - \frac{\sum_{i=1}^{n} (y_i - \hat{y}_i)^2}{\sum_{i=1}^{n} (y_i - \bar{y})^2}
\label{rsquared}
\end{equation}

where \(y_i\) is the manually counted ground truth, \(\hat{y}_i\) is the
predicted count from \dmc, and \(\bar{y}\) is the mean of the manually
counted ground truth.

\section{Results}
\label{sec:orga3eb9b0}

Our approach to seedling stand counting was to create a robust,
generalized detection model capable of handling various maize growth stages
(primarily V4--V8), planting methods, plant densities (including separated,
touching, and clustered plants), genetic lines, soil types, \uav altitudes,
flight trajectories, and camera poses.
Seedlings smaller than V2 consistently posed challenges due to insufficient
visual differentiation from weeds and background.
Larger seedlings beyond
V8 often overlapped, complicating accurate individual detections,
especially when clustered.
Balancing these trade-offs required careful
tuning of detection strategies, segmentation methods, and mosaic
processing.
We evaluated three main seedling counting pipelines using a range of input
modes and processing strategies:

\subsection{Color Segmentation and Voronoi-Based Stand Counting}
\label{sec:org1c8b93d}

Color-based segmentation was performed using the \exg index
applied to full-resolution mosaics.
After segmentation, we applied the distance transform and watershed to separate
clustered regions and generate individual Voronoi cells.
Each cell was treated as a distinct plant detected.
\begin{figure}[ht]
\centering
\includegraphics[width=1\linewidth]{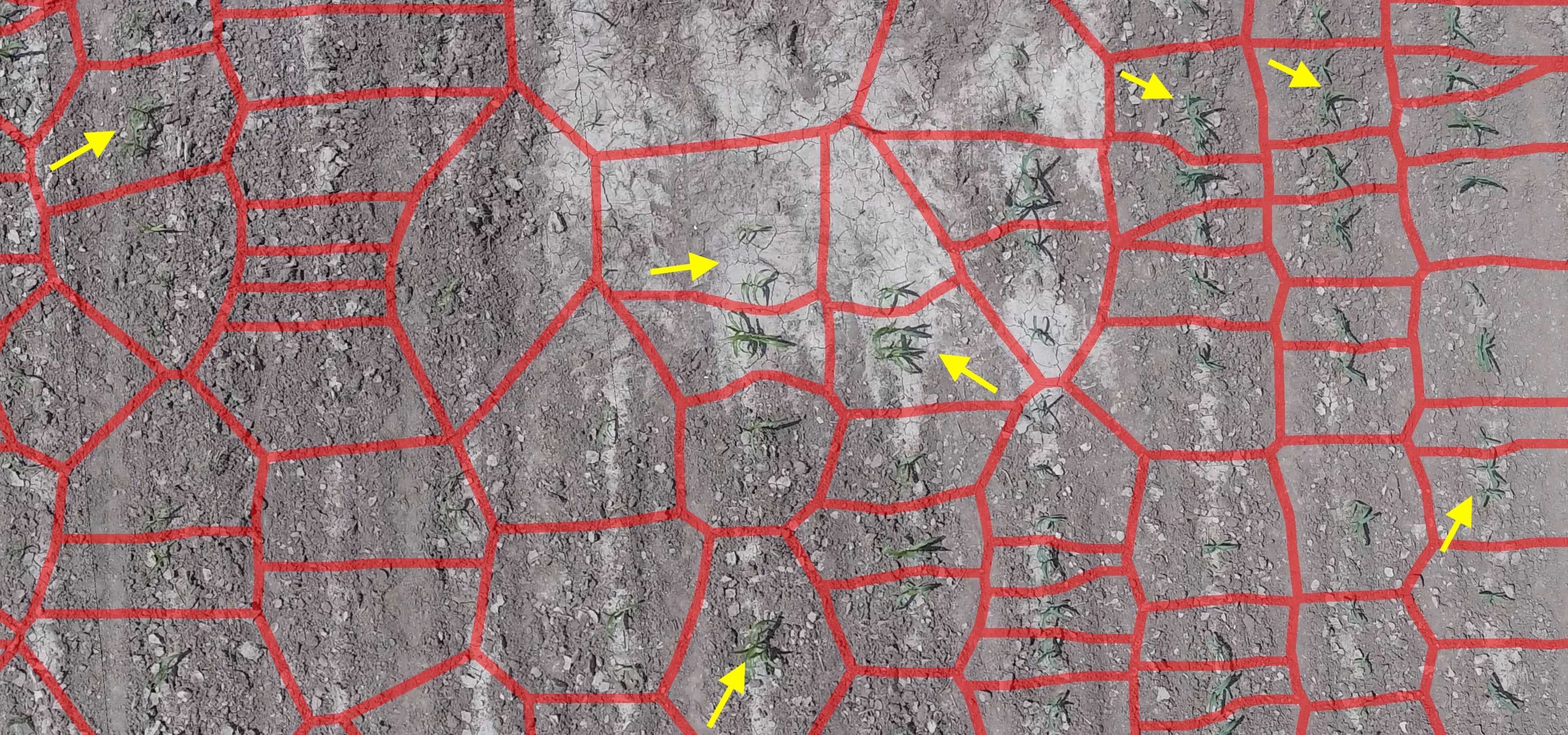}
\caption{Color-based segmentation under ideal field conditions. The field
section contains only maize, with minimal weed interference and
well-separated plants. Red lines denote Voronoi boundaries,
and yellow arrows indicate multiple plants within a single cell.}
\label{exg-nice}

\end{figure}
Figure \ref{exg-nice} presents the output of this segmentation process on a
section of a 2019 nursery field mosaic.
In this example, plant spacing is regular and soil contrast is high,
allowing the Voronoi-based method to segment the plants.
However, in some cases (indicated with yellow arrows), multiple well-separated
plants were not correctly distinguished and were marked as a single plant.

A more challenging test involved segmenting an entire field, where
additional green objects such as weeds and grasses were present alongside
the maize plants.
The mosaicking for this dataset was performed using \webodm
\cite{webodm2024}. Although some portions of the field were missing in the
final mosaic, all rows were fully captured, allowing for complete analysis.
%
%
The Radon transform is applied to the
\exg image to estimate the row orientation angle, and the image is then
rotated accordingly.
A binary mask is generated, and after applying the
distance transform and watershed algorithm, the foreground plant regions
are highlighted as teal blobs. Background boundaries, marked with red
lines, define the Voronoi cell borders used in the final segmentation
output.
%

\begin{figure}[ht]
\centering
\includegraphics[width=0.97\linewidth]{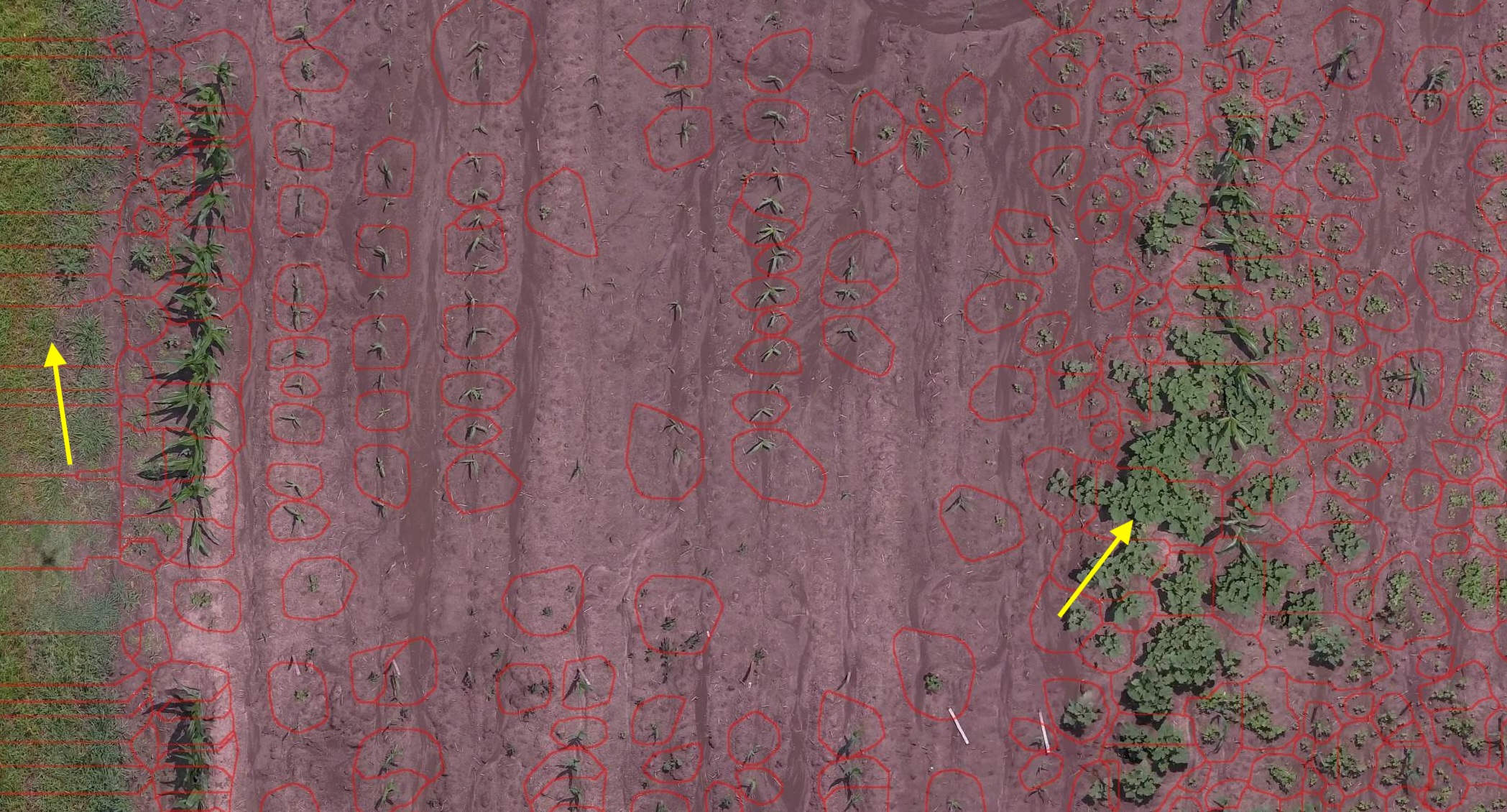}
\caption{Close-up view of the 2020 dataset showing maize seedlings interspersed with weeds and grass.}
\label{exg-weed}
\end{figure}

%
%
The algorithm performs well on well-separated maize plants, particularly in
the V3–V6 growth stages.
However, for larger plants such as those near the field borders, it
struggles. In some cases, a single plant is erroneously split into multiple
objects.
A closer look in Figure \ref{exg-weed} highlights the challenge of
distinguishing maize seedlings from non-crop vegetation in color-based
segmentation.
This highlights that the method segments all green vegetation
indiscriminately, regardless of whether it is maize or not.

\subsection{\dmc on Mosaic Images}
\label{sec:orga7d3054}

\begin{figure}
\centering
\includegraphics[width=0.96\linewidth]{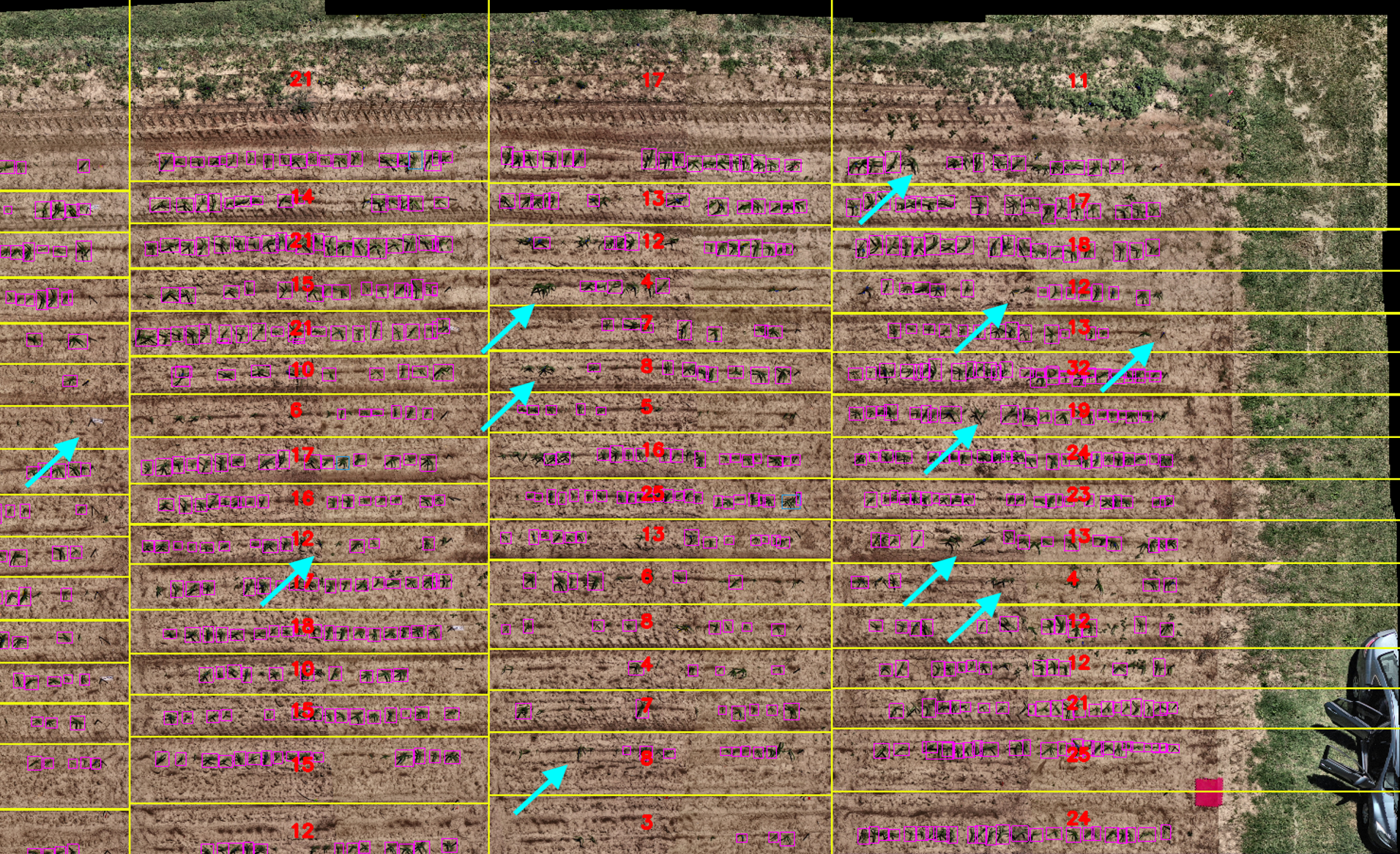}
\caption{A snippet of the final stand count of \dmc in mosaic mode.
Detected plants are shown with bounding boxes:
magenta for single plants, blue for doubles, and green for triples.
Range and row boundaries are indicated with yellow lines,
while missing detections are marked with blue arrows.
}
\label{final-moz}
\end{figure}

For \yolo-based detection on mosaicked images, the mosaics were first
divided into overlapping fragments (typically 1280\texttimes{}1280 pixels
with 10\% overlap; these values can be customized by the user) to
meet the input size constraints of the network.
Predictions from overlapping regions were merged using NMS to remove duplicate detections.
%
%
%
Panel \ref{final-moz} demonstrates the result of range and row detection,
along with seedling counts per row. Yellow lines indicate the detected
range and row boundaries, and the number of detected seedlings per row is
displayed in red at the center of each row.
Detection performance was strong in well-separated, clean regions, where
most maize seedlings were successfully identified. The model showed good
ability to distinguish maize from surrounding weeds. However, large
seedlings that were touching or tightly clustered were occasionally
under-detected, likely due to overlapping shapes or insufficient separation
cues in the visual signal.
In addition, some isolated plants were missed altogether, often when their
appearance was atypical or their color contrast with the soil background
was weak, leading the model to confuse them with background.


\begin{figure}[!t]
\centering
    \includegraphics[width=1\linewidth]{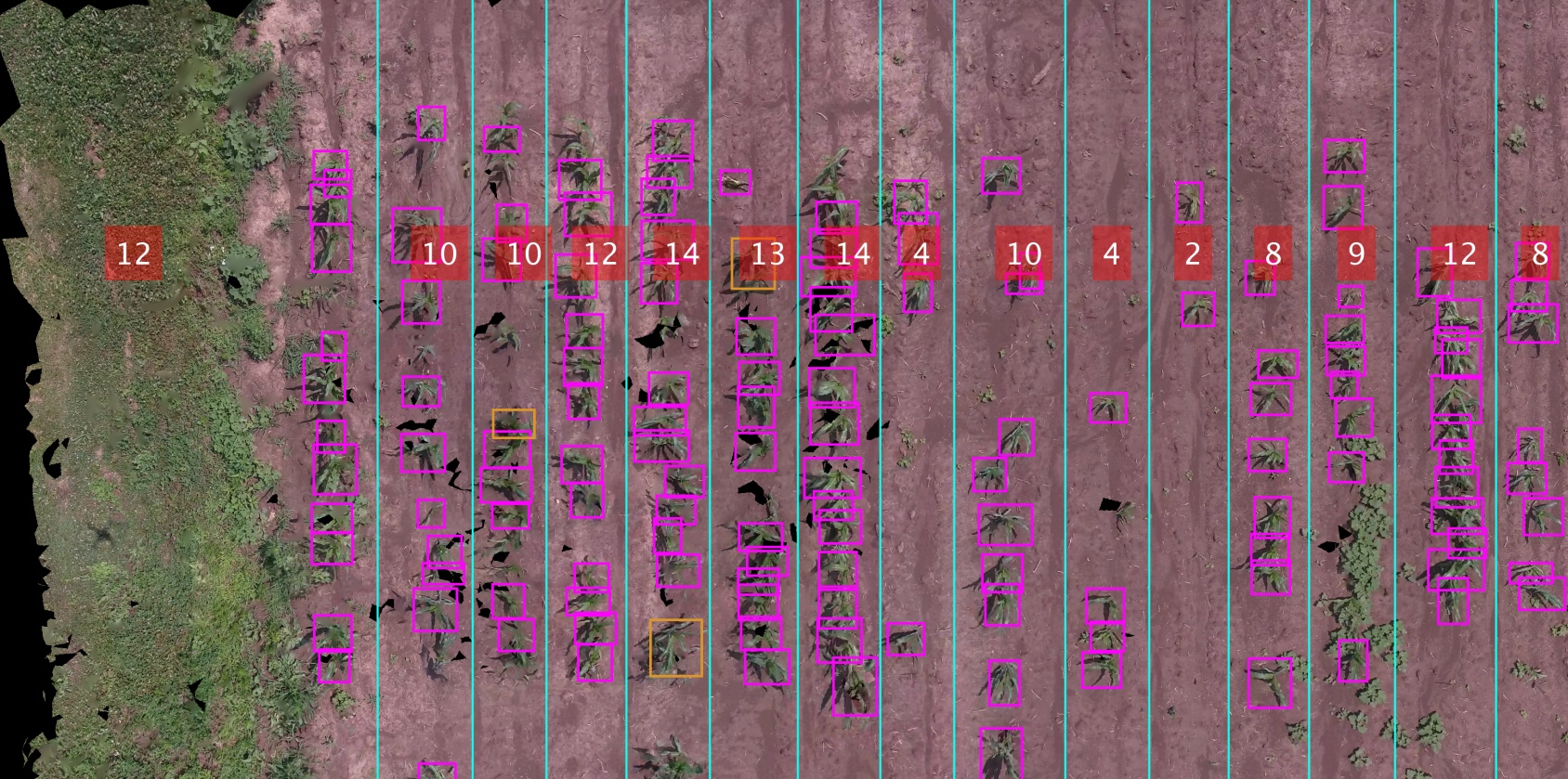}
\caption{\masc is robust to the poorer mosaics generated by \webodm. 
The mosaic exhibits artifacts such as holes (black pixels) and misregistered “melted” maize seedlings. 
Despite these issues, \masc in mosaic mode detected and counted most seedlings accurately.}
\label{webodm-det}
\end{figure}

We also performed stand counting on a mosaic generated using
\webodm.
However, its quality was noticeably lower compared to mosaics
produced by \cnvt and \cnvtt.
As shown in Figure \ref{webodm-det},
the \webodm mosaic exhibits visible artifacts, such as holes (black
pixels) and misregistered maize seedlings that appear distorted or
``melted''.
Despite these issues, \yolo was still able to detect most
seedlings effectively, with only a small number of missed detections.

\subsection{\dmc on Raw Video Frames}
\label{sec:org817f437}

\begin{figure*}
\centering
\includegraphics[width=0.96\linewidth]{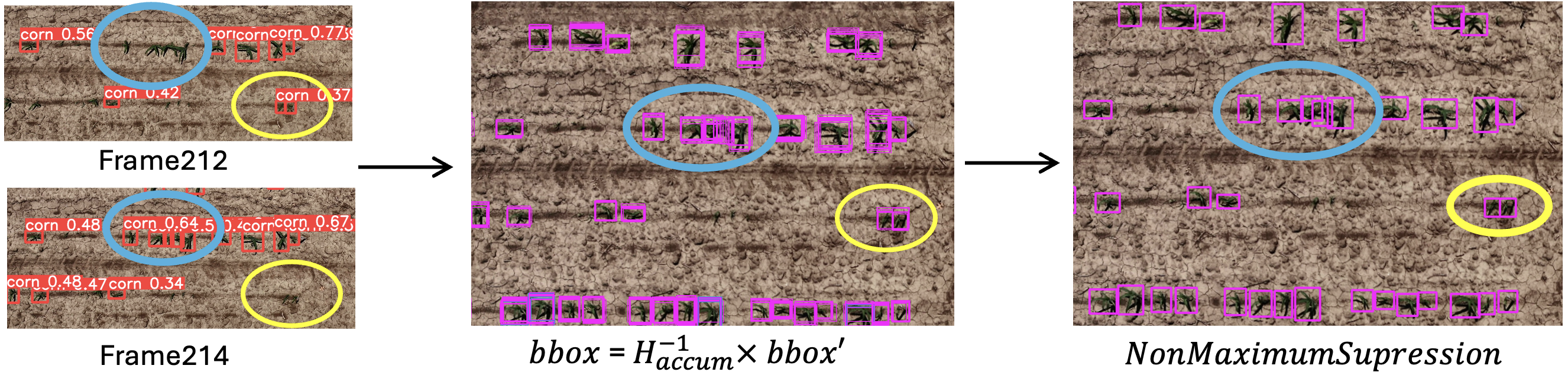}
\caption{Using raw video frames resolves missing and replicated plants.
Left: Two frames showing missed and recovered detections.
The plants in the blue circle were missed in Frame212 but detected in Frame214.
The plants in the yellow circle were detected in Frame212 but missed in Frame214.  
Middle: Aggregated bounding boxes before consensus.  
Right: Final result after applying Non-Maximum Suppression (NMS) to keep the strongest bounding box.}
\label{raw-det}
\end{figure*}

\begin{figure}
\centering
\includegraphics[width=0.96\linewidth]{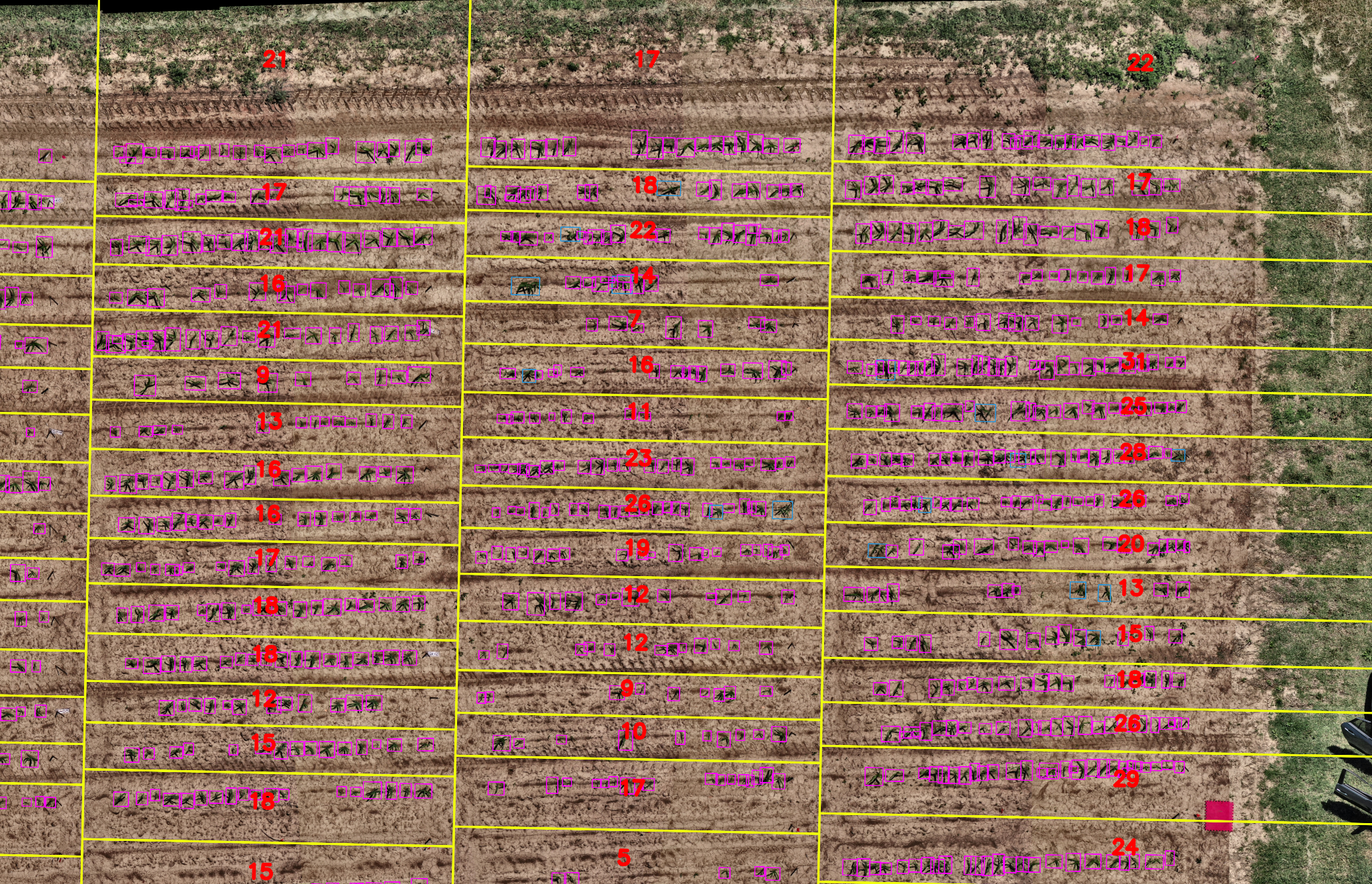}
\caption{A snippet of the final stand count of \dmc in raw frame mode.
Detected plants are shown
with bounding boxes:
magenta for single plants, blue for doubles, and green for triples.
Range and row boundaries are indicated with yellow lines.}
\label{raw-final}
\end{figure}

In the raw video mode, the input video is processed by \cnvt, which
performs dynamic frame sampling, camera and gimbal calibration, homography
estimation, shot detection, and mini-mosaicking. These steps result in the
construction of a global mosaic.
Each calibrated frame is then passed through the \yvn network, producing
bounding boxes and confidence value for each detected object.
Using the homography matrices computed during mosaicking, all bounding
boxes are projected onto the global coordinate system, then aggregated and
recorded in a single final label file.
Due to the high overlap between frames, individual plants often appear in
multiple frames.
To eliminate duplicate detections, NMS is applied
with both the NMS and confidence score thresholds set to 25\%. Only the
bounding box with the highest confidence is retained for each overlapping
instance.
The result is a unified global mosaic annotated with filtered bounding
boxes for each plant.
Figure \ref{raw-det} illustrates this workflow, from individual frame-level
detections to the final globally aligned composite.
The same plant, detected across multiple frames, is consolidated into a
single box with the highest confidence score.
The likelihood of successful detection increases under this approach, as
plants missed in one frame are often captured in subsequent frames.
The blue and yellow circles in Figure \ref{raw-det} highlight this effect:
\yvn misses the plants inside the blue circle in
\texttt{Frame212},
but successfully detects them in
\texttt{Frame214} (Figure \ref{raw-det}), as well as in \texttt{Frame213} and
\texttt{Frame215} (data not shown).
Finally, range and row detection, along with per-row counting, is conducted
using the same procedure as in the mosaic mode pipeline.
The complete result is presented in Figure \ref{raw-final}.

\subsection{Evaluation}
\label{sec:org93593ad}

\begin{figure}[!t]
\centering
\subfloat[\masc mosaic\label{moz-r2}]{
    \includegraphics[width=0.47\linewidth]{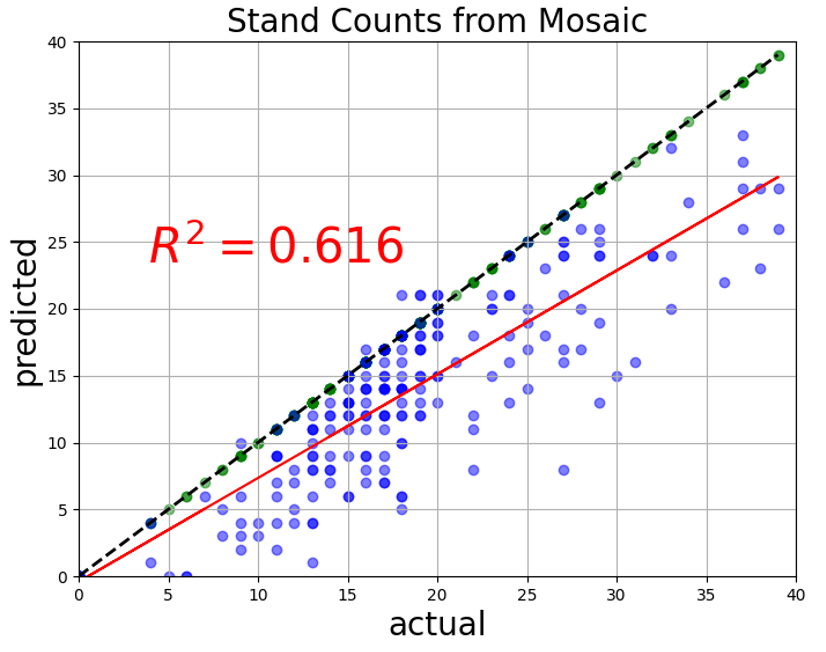}
}\hfil
\subfloat[\masc raw\label{raw-r2}]{
    \includegraphics[width=0.47\linewidth]{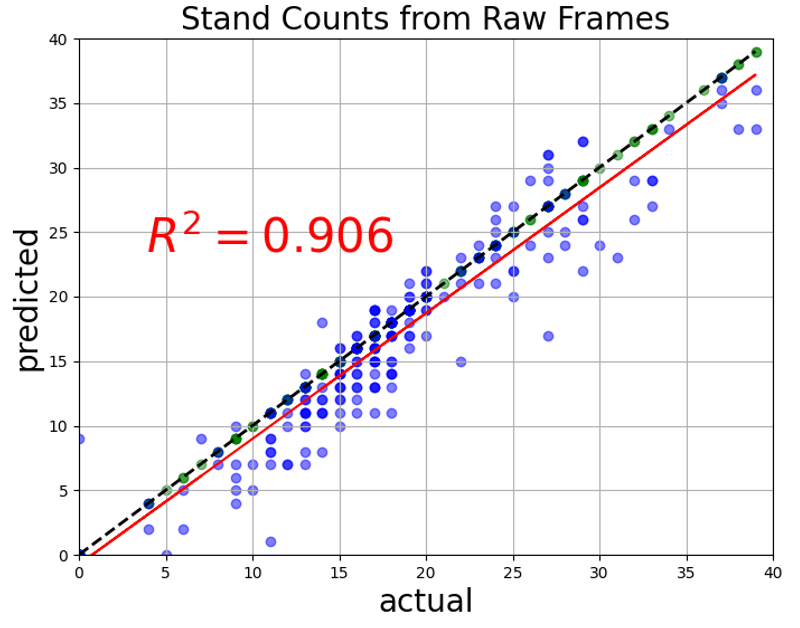}
}
\caption{$R^2$ evaluation results comparing mosaic mode and raw frame mode for stand counting. 
The mosaic-based approach achieved $R^2 = 0.616$, while the raw frame approach achieved $R^2 = 0.906$. 
The red line indicates the regression fit between predicted and actual counts, and the dashed line represents the ideal 1:1 relationship.
}
\label{eval-r2}
\end{figure}

To evaluate counting accuracy, we compared the per-row stand counts
produced by \dmc using both mosaic and raw video modes against manual
ground truth.
Figure \ref{eval-r2} presents the coefficient of determination (\(R^2\)) for
each approach.
The mosaic-based method achieved an \(R^2\) of only 0.616, indicating
moderate agreement but with noticeable deviations from the true counts.
In contrast, the raw frame-based pipeline yielded a significantly higher
\(R^2\) of 0.906, demonstrating strong linear correlation with manual counts.
This result suggests that processing from raw video frames, combined with
global bounding box aggregation and NMS, improves
accuracy, likely due to better recovery of missed detections and reduced
sensitivity to mosaicking artifacts.

\begin{figure}[hbtp]
\centering
\includegraphics[width=1\linewidth]{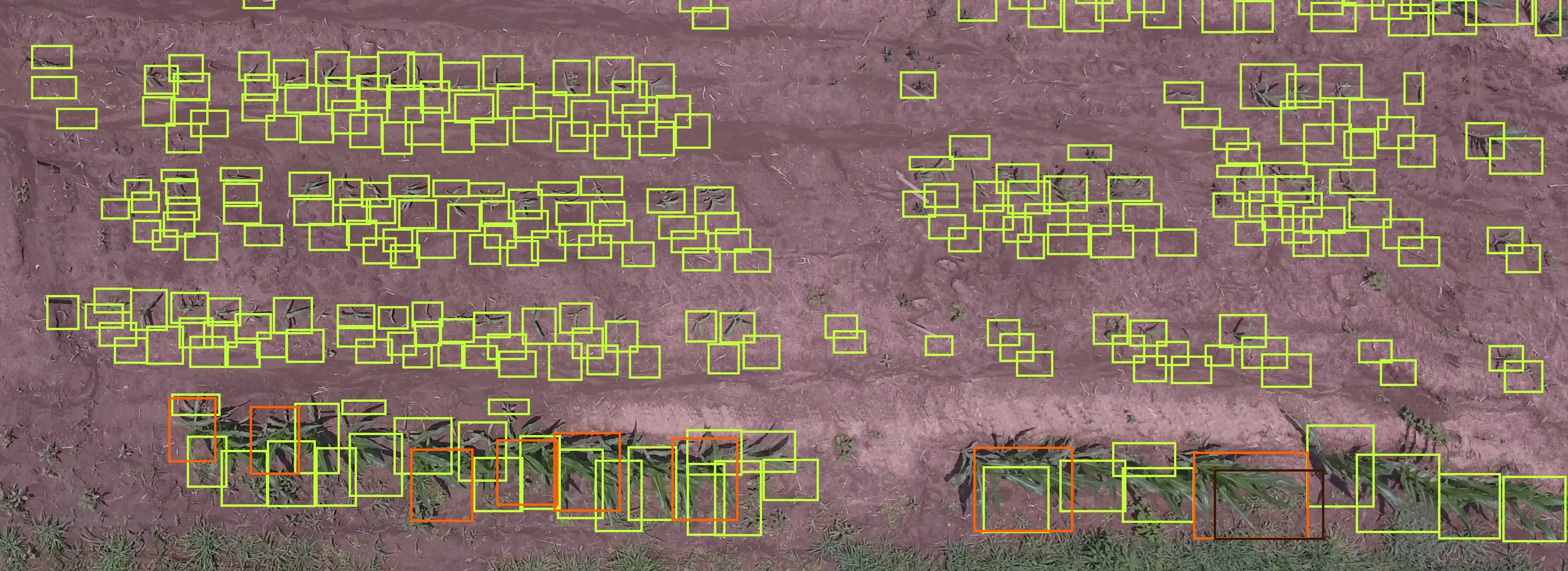}
\caption{Inaccurate homography matrices and poor-quality mosaics derived
from raw frame detection can degrade the accuracy of stand counting.
Misalignment between frames leads to improperly overlapped detections,
reducing the effectiveness of bounding box aggregation and increasing
counting errors.
}
\label{bad-raw}
\end{figure}

This significant improvement is largely due to the high quality of our
previous work in accurate homography estimation and the mosaicking
pipeline \cite{kharismawati2025a}.
When the homography matrix and mosaicking process are precise, multiple
bounding boxes for each plant align well and overlap correctly (see Figure
\ref{raw-det}, middle panel).
This enables reliable consensus based on the highest confidence scores.
In contrast, if the alignment is inaccurate, the bounding boxes may not
overlap as intended, leading to redundant detections that are not
eliminated.
This, in turn, negatively impacts evaluation performance.
Figure \ref{bad-raw} shows a sample case from our earlier \cn development,
where the homography estimation was not sufficiently robust
\cite{kharismawati2020}.

\section{Discussion}
\label{sec:org8b5716e}

Early season maize stand counting from aerial imagery presents numerous
challenges, stemming from biological, environmental, and technical
variabilities.
The first challenge we must address is distinguishing maize seedlings from
non-crop objects such as field weeds and grasses.
Our \exg-based segmentation methods are particularly limited in these
conditions.
This approach classifies all green pixels as vegetation, regardless of
their structural or contextual cues.
As a result, anything green --- including weeds, grasses, cover crops, or other
crops --- is detected as a seedling.
This simplistic reliance on pixel-level greenness, without incorporating
shape or pattern recognition, frequently results in overestimation of maize
counts in weedy or heterogeneous fields.
Figure \ref{exg-weed} illustrates this issue, where non-crop greenery is
misclassified as maize, highlighting the need for more robust
classification strategies.
In our field trials, plants were generally spaced at an average of 30
centimeters apart, though in some cases, spacing was as tight as 15
centimeters.
While tighter spacing helps compensate for poor germination rates,
it can also increase the visual complexity of separating individual plants,
especially as seedlings mature and begin to touch or overlap.
Color-based segmentation performed poorly when separating closely spaced
plants, even though the distance transform and watershed algorithms were
employed.
However, it still worked reasonably well on plants that were
well-separated.
Compared to \exg-based segmentation, our deep learning approach using the
\yolon model proved more robust.
Unlike \exg, which relies solely on greenness, \yolon leverages spatial and
morphological features to differentiate maize from other vegetation, making
it particularly effective in complex environments.
However, this is due to the class imbalance in the training
data.
Over 92\% of the annotated objects are single plants. Doubles and triples
are uncommon and usually result from planting errors, such as multiple
seeds being dropped into the same hole.
This imbalance can reduce detection accuracy for multi-plant clusters, but
it has little impact on final stand counts.
For example, if a group of plants is labeled as a triple-plant but detected
as a single and a double, it still contributes correctly to the total count
for that row, since our ground truth is based on row-level consensus.
A critical design decision lies in the trade-off between mosaic-based and
raw frame based detection.
In mosaic-based counting, mosaics must be split into overlapping fragments
because the \yolo architecture only processes inputs of size \(640 \times
640\) pixels.
Direct downsampling of mosaics was not viable, as our mosaic images can
reach sizes of \(5000 \times 10000\) pixels. Resizing them to \(640 \times
640\) would result in excessive loss of resolution, and forcing them into a
square aspect ratio would distort plant morphology, both of which
compromise detection accuracy.
Our patchification strategy maintained visual integrity and enabled
accurate detection without excessive memory use.
This mode yields a lower \(R^2\) value due to several contributing factors.
Mosaics can introduce visual artifacts from
stitching and blending, especially when rows span multiple frames with
imperfect registration.
Our use of alpha blending (\(\alpha = 1\)) exacerbates this, as pixels from
later frames replace earlier ones, often distorting seedling
shapes.
Furthermore, inaccuracies in the mosaic, such as errors in homography
estimation or insufficient image information in certain areas, can
introduce blurred or distorted regions, making seedling detection more
difficult.
This was evident in the mosaic generated with \webodm, where we observed
missing pixels and seedling artifacts caused by blending and interpolation,
often resulting in distorted or "melted" plant appearances.
Nevertheless,
\yolo successfully detected the vast majority of seedlings, overlooking
only a few.

The raw frame modes mitigates these issues by detecting objects in
calibrated, unstitched frames and then projecting bounding boxes onto a
global coordinate.
Each plant typically appears in multiple
overlapping frames, increasing the chance of successful detection.
While this approach is more computationally demanding --- it processes more
frames and full-sized, non-square images --- the \yolon model handled aspect
ratio changes well. Resizing from \(3807 \times 2073\) to \(640 \times 640\)
did not noticeably degrade detection quality.
Additionally, although each plant may generate multiple detections, NMS
effectively merges duplicates with minimal performance cost.
Edge ablation remains a challenge in raw frame detection. Partial plants
near the edges of frames may be incompletely captured, requiring careful
spatial aggregation across adjacent frames.
Nonetheless, the redundancy in raw frame input increases recall and
improves detection consistency.
Our quantitative evaluation supports the superiority of the raw frame
pipeline.
As shown in Figure \ref{eval-r2}, mosaic-based detection achieved an \(R^2\)
of 0.616, while raw frame based detection reached 0.906 -- demonstrating
strong linear agreement with manual counts.
This almost 30\% improvement highlights how the accurate homography
estimation and mosaicking pipeline developed in our previous work in
\cite{kharismawati2025a} directly enhances \dmc performance.
When alignment is precise, the bounding boxes for each plant line up well,
making it easy for NMS to reach a clear decision.
But when alignment is off, the boxes do not overlap properly, leading to
duplicate detections and worse evaluation scores.
These results clearly demonstrate that the methods introduced in this work
make a real, measurable difference in the overall accuracy of the
stand-counting pipeline.

One source of error that remains is the temporal mismatch between UAV
imagery and manual ground truthing.
In the 2024 season, imagery was collected seven days before manual counts.
During that time, some seedlings may have died (particularly for
fragile mutant lines), while others may have emerged after the aerial
survey.
These discrepancies likely influenced the \(R^2\) values.
Aligning imagery and ground truth collection to the same day will help
eliminate this variability in future studies.
Nadir view imagery also limits visibility of seedlings obscured by larger
neighbors.
Small or late-emerging plants hidden under the canopy cover may not be visible
from directly above.
Oblique imagery could help capture stem features, but introduces scale
distortion: seedlings closer to the camera appear disproportionately large
compared to those farther away.
While the \yolon model can detect from oblique angles, its performance
declines at the image edges where distortion is greatest.

\section{Conclusion}
\label{sec:org9a4de06}

Despite the challenges of weed interference, class imbalance, temporal
mismatches, and occlusion effects, our deep learning-based detection
pipeline shows strong potential for robust maize stand counting.
The compact and efficient \yolon model achieved near real-time inference
while maintaining high accuracy, particularly in raw frame mode, which
reached an \(R^2\) of 0.906 against manual counts.
Looking forward, synchronizing manual and image-based counts, improving
crop \emph{vs} weed classification, addressing class imbalance in training
data, and exploring oblique or multi-angle imagery will be important next
steps to improve accuracy.
Overall, our results indicate that deep learning applied to raw \uav video
frames offers a scalable, accurate, and resilient solution for maize stand
counting across diverse and complex field conditions.

\subsection*{Acknowledgments}
\label{sec:orgb6618e8}
We gratefully acknowledge our colleagues at the Missouri Maize Center for their
collaboration and support, particularly our farm manager
Chris Browne for his dedicated efforts in field operations.
We also extend our thanks to Filiz Bunyak, Hadi AliAkbarpour, Kannappan Palaniappan, Matt
Stanley, Dexa Akbar, Rifki Akbar, Bill Wise, and Vinny Kazic-Wise for their
thoughtful input and ongoing support during the development of this work.

\subsection*{Funding}
\label{sec:org3a7d00f}
We sincerely thank the Department of Electrical Engineering and Computer
Science for supporting D. Kharismawati, as well as an anonymous donor for their
generous contribution toward maize research.

\subsection*{Conflict of Interest Statement}
\label{sec:orgfbbd80b}
The authors declare that the research was conducted in the absence of any
commercial or financial relationships that could be construed as a
potential conflict of interest.

\subsection*{Data Availability}
\label{sec:org250545c}
The seedling maize detection dataset is available at  \href{https://drive.google.com/drive/folders/1EC2aR1HbCsRnIJbXWeEHddyGTQpGFh2n?usp=sharing}{\smd dataset} \cite{kharismawati2025e}.
The raw frames and homography matrices from the 2024 \dmc raw frame
experiment are publicly available at \href{https://drive.google.com/drive/folders/1Sp7bnrQY9pNwTgajMRI-r-ceZCcV5Q2E?usp=sharing}{\dmc dataset}.
The source code for color-based segmentation, mosaic mode, and raw frame mode is
available at \href{https://github.com/dek8v5/MaizeStandCounting\_MaSC.git}{\dmc Github Repository}.

\bibliographystyle{IEEEtran}
\def\localdots{../../..}

%
%

%
%


\def\db#1{\localdots/bibliography/#1.bib}




%



%
 
\def\bp{\db}

%

\bibliography{\bp{journals},%
              \bp{keys},%
              \bp{miscellaneous},%
              \bp{clean-egbib},%
              \bp{nascent},%
              \bp{all}}

\begin{IEEEbiography}[
{\includegraphics[width=1in,height=1.25in,clip,keepaspectratio]{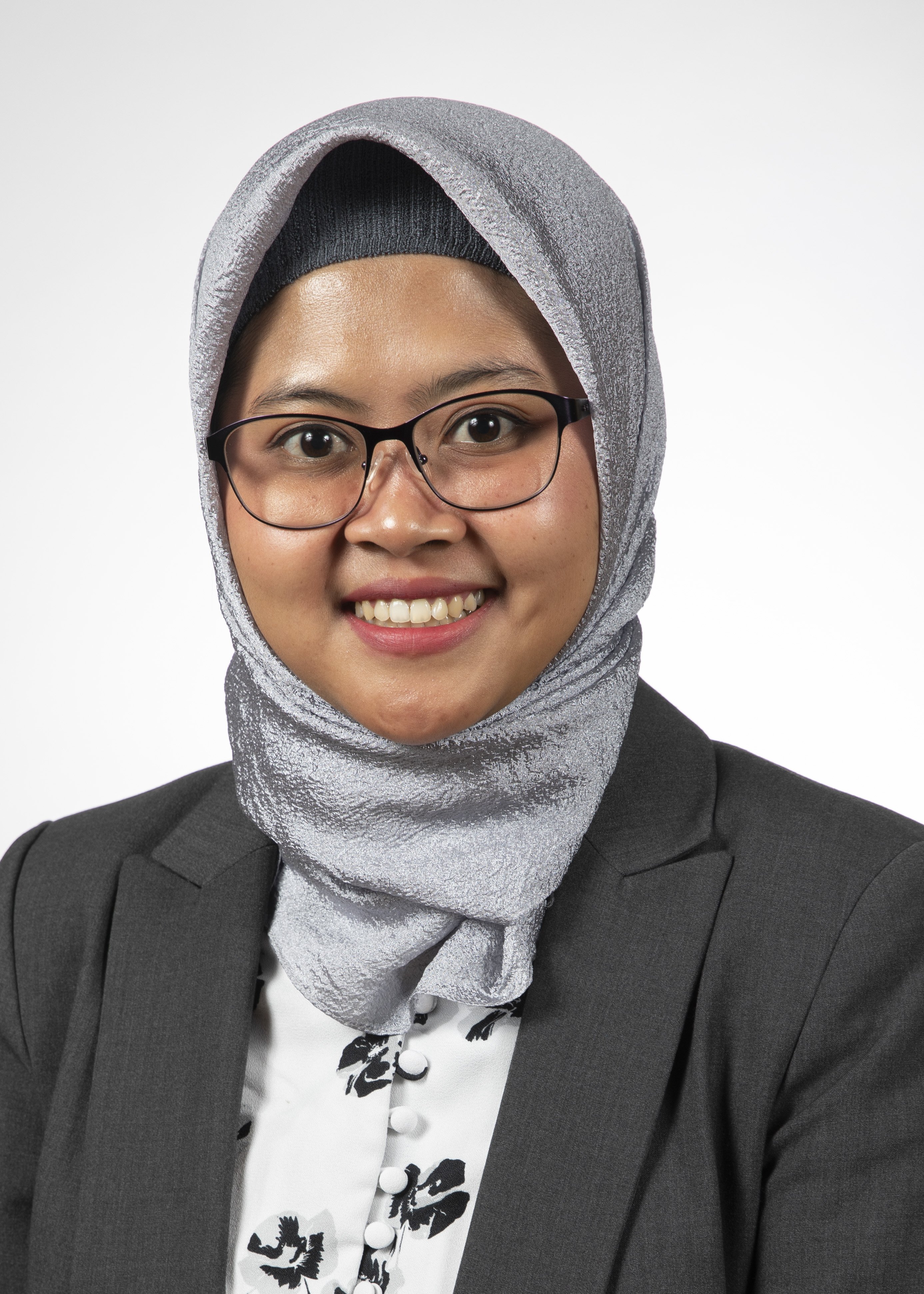}}]
{Dewi Endah Kharismawati} (Member, IEEE)
received the B.S. and Ph.D. degrees in Computer Science from the University of Missouri, Missouri, USA,
in 2017 and 2025, respectively. She has been a Postdoctoral Researcher with the AgSensing Lab, Department
of Food, Agricultural, and Biological Engineering, The Ohio State University, since June 2025.
Her research interests include computer vision, deep learning, and image processing applied to
UAV aerial imagery, with a focus on image registration, plant detection, and 3D reconstruction for
generating actionable insights for farmers and agricultural researchers.
\end{IEEEbiography}
\begin{IEEEbiography}[
{\includegraphics[width=1in,height=1.25in,clip,keepaspectratio]{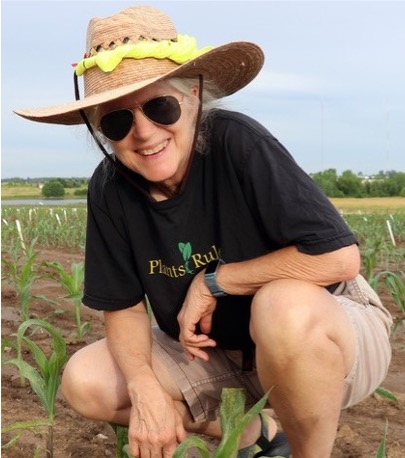}}]
{Toni Kazic} received the B.S. degree in microbiology from the University of Illinois,
Urbana, in 1975, the Ph.D. degree in genetics from the University of Pennsylvania in 1984,
and was a postdoctoral fellow in bacterial genetics at the Institute for Cancer Research,
Fox Chase Cancer Center, Philadelphia, and Washington University, St. Louis before switching
to computational biology.
Since 2006, her work has focused on computational and genetic studies of the
biochemical networks underlying complex phenotypes in maize. She is an associate
professor in the Department of Electrical Engineering and Computer Science at the
University of Missouri, Columbia. Her lab, the Missouri Maize Computation and Vision
(MMCV) Lab, investigates a family of maize mutants that exhibit lesion phenotypes,
which are visible spots on leaves and define a high-dimensional phenotypic manifold.
To scale up field experiments for higher resolution, better sample size, and more robust
quantitation, her group combines high throughput phenotyping with consumer-grade drones
with developing computational methods to monitor plant growth and lesion formation.
She has been a fellow at the NIH (Division of Computer Research and Technology),
Argonne National Laboratory (Division of Mathematics and Computer Science), and ICOT,
the Japanese Institute for Fifth Generation Computer Technology.
She has served as a Program Director at the National Science Foundation in
computational biology, as board member and secretary of the International Society
of Computational Biology, and as an associate member of the Joint Nomenclature Commission
of the International Union of Pure and Applied Chemistry and the International Union of
Biochemistry and Molecular Biology (the ``Enzyme Commission'').
She was elected a fellow of the American College of Medical Informatics in 2004.
\end{IEEEbiography}

\end{document}